\documentclass[twoside,11pt]{article} 
\usepackage{jmlr2e} 
\usepackage[utf8]{inputenc}

\usepackage{amsmath,amssymb,amsfonts,dsfont}  
\usepackage{url}
\usepackage[svgnames]{xcolor}
\usepackage{booktabs}
\usepackage[boxed,vlined]{algorithm2e}
\usepackage{listings}

\newcommand{\toolsuitename}{BOAH}

\usepackage{tikz}
\usetikzlibrary{shapes.geometric}
\usetikzlibrary{positioning,shapes,shadows,arrows,calc}
\usetikzlibrary{intersections}
\pgfdeclarelayer{background}
\pgfdeclarelayer{foreground}
\pgfsetlayers{background,main,foreground}

\tikzstyle{activity}=[rectangle, draw=black, rounded corners, text centered, text width=8em]
\tikzstyle{data}=[rectangle, draw=black, text centered, text width=8em]
\tikzstyle{myarrow}=[->, thick, draw=black]

\newcommand{\note}[1]{
	\noindent~\\
	\fcolorbox{Red}{Orange}{\parbox{0.99\textwidth}{#1}}
}
\renewcommand{\note}[1]{}

\renewcommand{\vec}[1]{\mathbf{#1}}

\RestyleAlgo{algoruled}
\DontPrintSemicolon
\LinesNumbered
\SetAlgoVlined
\SetFuncSty{textsc}

\SetKwInOut{Input}{Input}
\SetKwInOut{Output}{Output}
\SetKw{Return}{return}

\begin{document}

\title{\toolsuitename{}: A Tool Suite for Multi-Fidelity Bayesian Optimization \& Analysis of Hyperparameters}

\author{\name Marius Lindauer$^{1}$ \email lindauer@cs.uni-freiburg.de\\
\name Joshua Marben$^{1}$ \email marbenj@cs.uni-freiburg.de\\
\name Philipp Müller$^{1}$ \email muelleph@cs.uni-freiburg.de\\
\name Katharina Eggensperger$^{1}$ \email eggenspk@cs.uni-freiburg.de\\
\name Matthias Feurer$^{1}$ \email feurerm@cs.uni-freiburg.de\\
\name André Biedenkapp$^{1}$ \email biedenka@cs.uni-freiburg.de\\
\name Frank Hutter$^{1,2}$ \email fh@cs.uni-freiburg.de\\
\addr $^{1}$University of Freiburg, Germany\\
\addr $^{2}$Bosch Center for Artificial Intelligence, Germany}

\editor{TBD}

\maketitle

\begin{abstract}
Hyperparameter optimization and neural architecture search can become prohibitively expensive for regular black-box Bayesian optimization because the training and evaluation of a single model can easily take several hours. To overcome this, we introduce a comprehensive tool suite for effective multi-fidelity Bayesian optimization and the analysis of its runs. The suite, written in Python, provides a simple way to specify complex design spaces, a robust and efficient combination of Bayesian optimization and HyperBand, and a comprehensive analysis of the optimization process and its outcomes.
\end{abstract}

\section{Introduction}

Finding well-performing hyperparameter settings for a machine learning method often makes the difference
between achieving state-of-the-art or quite weak performance. While many prominent methods, such as Bayesian optimization (BO, \citeauthor{shahriari-ieee16a} \citeyear{shahriari-ieee16a}) or random search~\citep{bergstra-jmlr12a} formulate hyperparameter tuning as a blackbox optimization problem, the long training time and the demand
for large computational power of contemporary machine learning methods, such as deep neural
networks, limit the usefulness of these methods: when single function evaluations require days or
weeks, blackbox optimization becomes computationally infeasible. Recent advances in
hyperparameter optimization therefore go beyond this limiting blackbox formulation and consider
cheap, approximate function evaluations (a.k.a. evaluation budgets), such as performance when running on a subset of data, optimizing a deep neural network for only a few epochs or down-sampled images in computer vision (cf \citeauthor{feurer-automlbook18a} \citeyear{feurer-automlbook18a}). 

Here, we present the first comprehensive tool suite for such multi-fidelity optimization, called \toolsuitename{}, allowing users to not only optimize their hyperparameters much more effectively, but to also automatically analyze the optimization process and the importance of the various hyperparameters. This automatic analysis provides much-needed insights into the algorithm being optimized that is not available from executing any other multi-fidelity hyperparameter optimization package we are aware of.

\section{Related Work}

While there are many available tools for standard BO, e.g., Spearmint \citep{snoek-NeurIPS12a}, SMAC~\citep{hutter-lion11a}, HyperOpt~\citep{bergstra-NeurIPS11a}, GPyOpt and BOTorch, the same cannot be said for multi-fidelity BO; we are only aware of RoBO~\citep{klein-bayesopt17}, BOHB~\citep{falkner-icml18a}, and Dragonfly~\citep{kandasamy-arxiv19a}. Out of these, RoBO and Dragonfly use Gaussian Processes (GPs), which scale cubically in the number of data points and are thus problematic for multi-fidelity settings in which it is possible to gather many cheap data points. GPs also require carefully chosen hyperpriors and typically struggle to perform well in high-dimensional, categorical design spaces. 
In contrast, BOHB uses kernel density estimators (KDEs), similar to Tree Parzen Estimators as used in HyperOpt~\citep{bergstra-NeurIPS11a}, which work well across a wide range of problems. We therefore include BOHB as part of the \toolsuitename{} tool suite; we also strive to integrate other state-of-the-art optimization tools into \toolsuitename.
We emphasize that all existing open-source hyperparameter optimization tools we are aware of only focus on yielding good performance, not on aiding intuition and scientific understanding by means of automated analyses. 

\section{The \toolsuitename{} Tool Suite}

\begin{figure}
    \centering
    \scalebox{0.80}{
    \begin{tikzpicture}[node distance=4cm, thick]
	\node (function) [data] {Function $f$};
	\node (budgets) [data, below of=function, node distance=1cm] {Set of budgets $b$};
	\node (space) [data, below of=budgets, node distance=1cm] {Design Space $\mathcal{X}$};
	
	\node (pcs) [activity, dashed, right of=space, xshift=-0.2cm] {ConfigSpace};
	\node (hb) [activity, right of=pcs, yshift=-0.0cm] {Hyperband iteration};
    \node (kde) [activity, above of=hb, node distance=2cm] {BO model};
	
    \draw[myarrow] ($(kde.south)+(-0.3,0.0)$) -- ++(0.0,-0.6) node[left] {$\vec{x} \in \mathcal{X}$} -- ($(hb.north)+(-0.3,+0.0)$);
	\draw[myarrow] ($(hb.north)+(0.3,+0.0)$) -- ++(0.0,0.6) node[right] {$f_b(\vec{x})$} -- ($(kde.south)+(0.3,0.0)$);
	
	\draw[myarrow] (function.east) -- ($(kde.west)+(-0.3,0.0)$);
	\draw[myarrow] (space.east) -- (pcs.west);
	\draw[myarrow] (budgets.east) -- ($(kde.west)+(-0.3,-1.)$);
	\draw[myarrow] (pcs.east) -- ($(kde.west)+(-0.3,-2.)$);
	
	\node (perf) [activity, right of=kde, node distance=6cm] {Performance Analysis};
	\node (budget) [activity, below of=perf, node distance=1cm] {Budget Analysis};
	\node (imp) [activity, below of=budget, node distance=1cm] {Hyperparameter Importance};
	
	\draw[myarrow] ($(kde.east)+(0.3,-1.)$) -- node[above] {$\langle \vec{x}_i, f_{b_i}(\vec{x}_i)\rangle_i$} ($(perf.west)+(-0.3,-1.)$);
	\draw[myarrow] ($(kde.east)+(0.3,-1.)$) -- node[below] {$\vec{x}^*$} ($(perf.west)+(-0.3,-1.)$);
	
	\begin{pgfonlayer}{background}
    
    	\path (kde -| kde.west)+(-0.25,0.85) node (resUL) {};
    	\path (hb.east |- hb.south)+(0.25,-0.5) node(resBR) {};
    	\path [rounded corners, draw=black!60, dashed] (resUL) rectangle (resBR);
		\path (hb.east |- hb.south)+(-.5,-.3) node [text=black!60] {BOHB};
    	
    	\path (perf -| perf.west)+(-0.25,0.85) node (resUL) {};
    	\path (imp.east |- imp.south)+(0.25,-0.5) node(resBR) {};
    	\path [rounded corners, draw=black!60, dashed] (resUL) rectangle (resBR);
		\path (imp.east |- imp.south)+(-.5,-.3) node [text=black!60] {CAVE};
    	
    \end{pgfonlayer}
	
\end{tikzpicture}
}
    \caption{Workflow of \toolsuitename{} }
    \label{fig:workflow}
\end{figure}
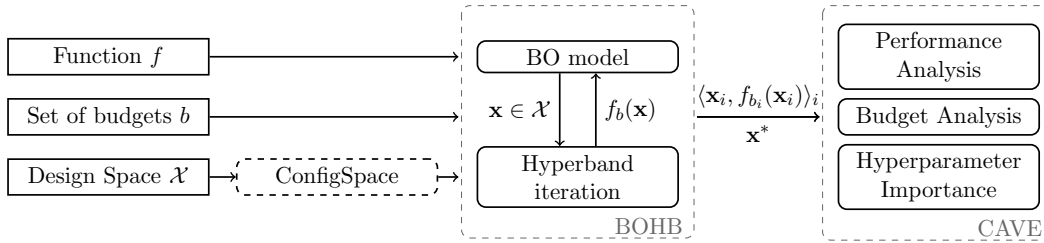

\toolsuitename{}\footnote{\url{https://www.automl.org/\toolsuitename}. We provide at \toolsuitename's website also a links to all packages and to a repository with many examples as Jupyter notebooks on how to combine these packages.} consists of three modular packages that interact as shown in the workflow in Figure~\ref{fig:workflow}. In the following, we describe the individual packages in turn.

\subsection{ConfigSpace: Definition of the Design Space}
\label{ssec:configspace}

To specify the design space $\mathcal{X}$, we provide a package dubbed \texttt{ConfigSpace}. It supports all common hyperparameter types, such as \emph{categorical}, \emph{ordinal}, \emph{integer-valued} and \emph{continuous} hyperparameters. Furthermore, users can define \emph{conditional constraints} between hyperparameters (e.g., hyperparameters of the RBF kernel are only active if 
a RBF kernel is chosen) or whether or not a continuous hyperparameter should be sampled on a logarithmic scale (such as the learning rate for gradient descent algorithms).
This package can also be used on its own and is already used by other AutoML packages, such as \texttt{SMAC3}~\citep{lindauer2017smac} or \texttt{Auto-Sklearn}~\citep{feurer-NeurIPS2015a}.

\subsection{BOHB: Bayesian Optimization with Hyperband}

As optimizers, \toolsuitename{} supports \texttt{BOHB}~\citep{falkner-icml18a} and its components, including successive halving \citep{jamieson-aistats16}, HyperBand~\citep{li-jmlr18a}, a hyperopt-like optimizer \citep{bergstra-NeurIPS11a} and the combination of BO and HyperBand.
The main idea of the latter is to use BO to suggest new configurations $\vec{x} \in \mathcal{X}$ and to efficiently determine the best $\vec{x}^*$ with the help of HyperBand evaluating $f_b(\vec{x})$ on different budgets $b$.

Due to its flexibility and strong anytime performance, BOHB is a well-suited optimizer for several settings, including DNN architectures with number of epochs being the budgets (e.g., Auto-PyTorch), hyperparameters of RL algorithms with episode length being the budgets and hyperparameters of machine learning algorithms (e.g., SVMs) with number of validated cross validation folds being the budgets. Another advantage of BOHB compared to other BO-tools is its efficient parallelization. 
To make BOHB as easy to use as possible, we added a new high-level \emph{fmin} interface, inspired by the well-known \emph{fmin} interface of scipy.   

\subsection{CAVE: Visualization of Results}
\label{ssec:cave}

For analyzing the data collected by BOHB, we integrated CAVE~\cite{biedenkapp-lion18a} into \toolsuitename.
To this end, we extended CAVE to be able to analyze multi-fidelity data and to help users to gain a better understanding of how BOHB searched a given design space. On each budget and across budgets, CAVE provides extended analyses of:

\begin{description}
    \item[Hyperparameter importance] using local parameter importance~\citep{biedenkapp-lion18a} and fANOVA~\citep{hutter-icml14a} to analyze which hyperparameters were the most important ones to achieve high performance. To be applicable to BOHB, CAVE provides analyses for each budget and adds uncertainty estimates for each budget.
    \item[Rank correlation between the budgets] to verify that the observations on the individual budgets are highly correlated. Multi-fidelity optimizers, such as BOHB, perform best if similar configurations perform best across the various budgets. Therefore, the degree to which this is the case is important to keep track of.
    \item[Optimizer footprint plots] to study how BOHB sampled the design space. For example, these plots help to identify general promising areas in the design space. CAVE uses multi-dimensional scaling to project the n-dimensional space into a 2-dimensional space under consideration of a distance metric in complex design spaces.
\end{description}  


\section{Show Case: Hyperparameter Optimization for Reinforcement Learning}

\note{ML: I discussed with André and Matze that we should replace that benchmark since the budget definition does not allow us to show some interesting results/insights. Nevertheless, we will submit the paper with the current benchmarks and will change it in the first review iteration.}

To showcase \toolsuitename{}'s flexibility and the usefulness of its automated analyses, we ran it on a rarely studied AutoML problem: hyperparameter optimization for reinforcement learning~\citep{falkner-icml18a}.\footnote{\url{https://github.com/automl/BOAH/blob/master/examples/icml_2018_experiments/cartpole.ipynb}} The goal is to obtain well-performing hyperparameters of Proximal Policy Optimization (PPO)~\citep{schulman-arxiv17a} such that PPO on average solves the cartpole problem within a minimal number of epochs. As budgets for BOHB, we used the number of PPO runs, since individual PPO runs provide a very noisy performance estimate and thus multiple repetitions yield a clearer signal. We ran \toolsuitename{} $10$ times for 128 iterations using $10$ workers on a compute cluster with nodes equipped with two Intel Xeon E5-2630v4 and $128$GB memory
running CentOS 7. We set the lowest budget to be $1$~repetition and allow $9$ repetitions for the highest budget.

\begin{figure}
    \centering
    \includegraphics[width=0.3\textwidth]{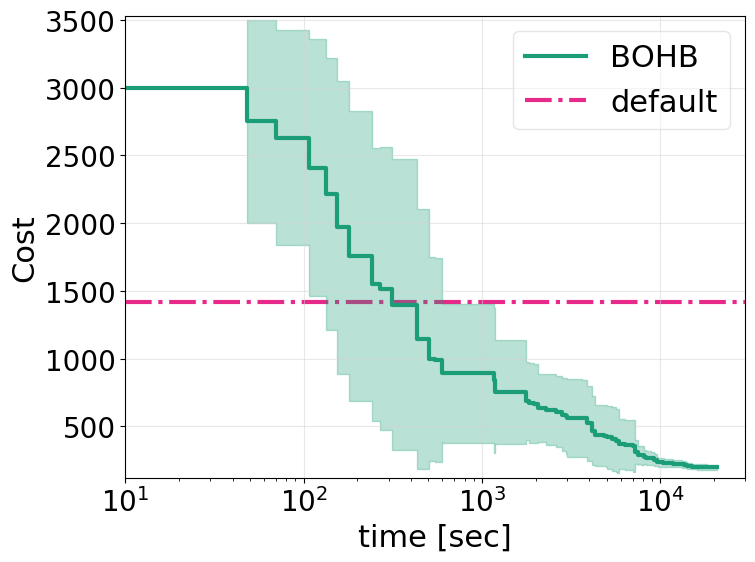}
    \includegraphics[width=0.37\textwidth]{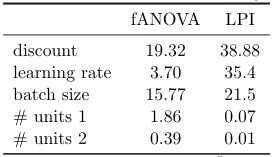}
    \includegraphics[width=0.31\textwidth]{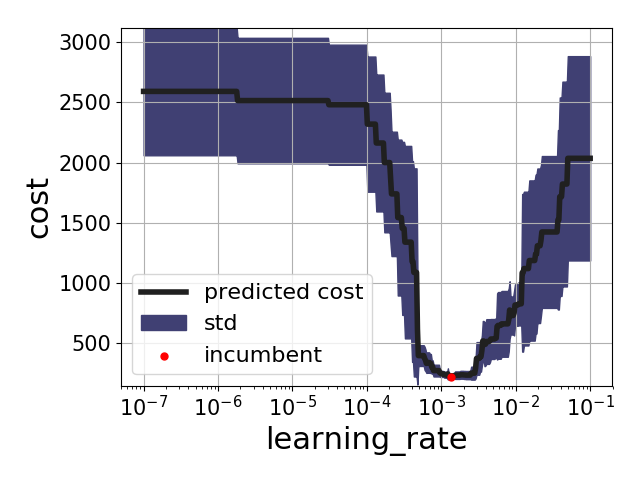}
    \caption{\emph{Left:} Performance of incumbent configuration found by BOHB over time. As a baseline we show the final cost of the PPO's default configuration. Cost refers to the number of epochs PPO needed to solve the cartpole problem.
    \emph{Middle:} Hyperparameter importance with fANOVA and LPI as percentages on the largest budget. Because of space limitations, we do not show uncertainty estimates. \emph{Right:} Estimated local hyper-parameter importance (LPI) analysis of the learning rate.}
    \label{fig:exp}
\end{figure}


The resulting CAVE report includes the following insights (Figure~\ref{fig:exp}):
(i) BOHB indeed improved the sample-efficiency of PPO substantially compared to the default settings of PPO;
(ii) according to the global analysis of the hyperparameter importance based on fANOVA, the discount and the batch size are the most important hyperparameters;
(iii) in addition to these, the learning rate is also very important according to the local hyperparameter importance analysis (LPI), also shown in the middle and right of Figure~\ref{fig:exp}.

\section{Conclusion}

We introduced \toolsuitename{}, a comprehensive suite of tools that allow users to conveniently specify design spaces, efficiently search these spaces using multi-fidelity Bayesian optimization, and analyze the results in a wide variety of ways. We emphasize that these steps are integrated to work together seamlessly, and that the posthoc analysis of the optimization process with CAVE does not require any additional function evaluations. To the best of our knowledge, \toolsuitename{} is the first tool suite which brings all these three important components together and therefore improves the usability of AutoML substantially.

\section*{Acknowledgments}
The authors acknowledge funding by the Robert Bosch GmbH, support by the state of Baden-Württemberg through bwHPC and the German Research Foundation (DFG) through grant no. INST 39/963-1 FUGG.

\newpage
\bibliography{strings,lib,local,shortproc}

\begin{thebibliography}{16}
\providecommand{\natexlab}[1]{#1}
\providecommand{\url}[1]{\texttt{#1}}
\expandafter\ifx\csname urlstyle\endcsname\relax
  \providecommand{\doi}[1]{doi: #1}\else
  \providecommand{\doi}{doi: \begingroup \urlstyle{rm}\Url}\fi

\bibitem[Bergstra and Bengio(2012)]{bergstra-jmlr12a}
J.~Bergstra and Y.~Bengio.
\newblock Random search for hyper-parameter optimization.
\newblock \emph{JMLR}, 13:\penalty0 281--305, 2012.

\bibitem[Bergstra et~al.(2011)Bergstra, Bardenet, Bengio, and
  K{\'e}gl]{bergstra-NeurIPS11a}
J.~Bergstra, R.~Bardenet, Y.~Bengio, and B.~K{\'e}gl.
\newblock Algorithms for hyper-parameter optimization.
\newblock In \emph{Proc. of NeurIPS'11}, pages 2546--2554, 2011.

\bibitem[Biedenkapp et~al.(2018)Biedenkapp, Marben, Lindauer, and
  Hutter]{biedenkapp-lion18a}
A.~Biedenkapp, J.~Marben, M.~Lindauer, and F.~Hutter.
\newblock Cave: Configuration assessment, visualization and evaluation.
\newblock In \emph{Proc. of LION'18}, 2018.

\bibitem[Falkner et~al.(2018)Falkner, Klein, and Hutter]{falkner-icml18a}
S.~Falkner, A.~Klein, and F.~Hutter.
\newblock {BOHB:} robust and efficient hyperparameter optimization at scale.
\newblock In \emph{Proc. of ICML}, pages 1436--1445, 2018.

\bibitem[Feurer and Hutter(2019)]{feurer-automlbook18a}
M.~Feurer and F.~Hutter.
\newblock Hyperparameter optimization.
\newblock In \emph{AutoML: Methods, Sytems, Challenges}, chapter~1, pages
  3--38. Springer, 2019.

\bibitem[Feurer et~al.(2015)Feurer, Klein, Eggensperger, Springenberg, Blum,
  and Hutter]{feurer-NeurIPS2015a}
M.~Feurer, A.~Klein, K.~Eggensperger, J.~T. Springenberg, M.~Blum, and
  F.~Hutter.
\newblock Efficient and robust automated machine learning.
\newblock In \emph{Proc. of NeurIPS'15}, pages 2962--2970, 2015.

\bibitem[Hutter et~al.(2011)Hutter, Hoos, and Leyton-Brown]{hutter-lion11a}
F.~Hutter, H.~Hoos, and K.~Leyton-Brown.
\newblock Sequential model-based optimization for general algorithm
  configuration.
\newblock In \emph{Proc. of LION'11}, pages 507--523, 2011.

\bibitem[Hutter et~al.(2014)Hutter, Hoos, and Leyton-Brown]{hutter-icml14a}
F.~Hutter, H.~Hoos, and K.~Leyton-Brown.
\newblock An efficient approach for assessing hyperparameter importance.
\newblock In \emph{Proc. of ICML'14}, pages 754--762, 2014.

\bibitem[Jamieson and Talwalkar(2016)]{jamieson-aistats16}
K.~Jamieson and A.~Talwalkar.
\newblock Non-stochastic best arm identification and hyperparameter
  optimization.
\newblock In \emph{Proc. of AISTATS'16}, 2016.

\bibitem[Kandasamy et~al.(2019)Kandasamy, Vysyaraju, Neiswanger, Paria,
  Collins, Schneider, Poczos, and Xing]{kandasamy-arxiv19a}
K.~Kandasamy, K.~Vysyaraju, W.~Neiswanger, B.~Paria, C.~Collins, J.~Schneider,
  B.~Poczos, and E.~Xing.
\newblock Tuning hyperparameters without grad students: Scalable and robust
  {B}ayesian optimisation with dragonfly.
\newblock \emph{arxiv:1903.06694}, 2019.

\bibitem[Klein et~al.(2017)Klein, Falkner, Mansur, and
  Hutter]{klein-bayesopt17}
A.~Klein, S.~Falkner, N.~Mansur, and F.~Hutter.
\newblock {RoBO}: A flexible and robust {B}ayesian optimization framework in
  {P}ython.
\newblock In \emph{NeurIPS Workshop: BayesOpt}, 2017.

\bibitem[Li et~al.(2018)Li, Jamieson, DeSalvo, Rostamizadeh, and
  Talwalkar]{li-jmlr18a}
L.~Li, K.~Jamieson, G.~DeSalvo, A.~Rostamizadeh, and A.~Talwalkar.
\newblock Hyperband: {A} novel bandit-based approach to hyperparameter
  optimization.
\newblock \emph{JMLR}, 18:\penalty0 185:1--185:52, 2018.

\bibitem[Lindauer et~al.(2017)Lindauer, Eggensperger, Feurer, Falkner,
  Biedenkapp, and Hutter]{lindauer2017smac}
M.~Lindauer, K.~Eggensperger, M.~Feurer, S.~Falkner, A.~Biedenkapp, and
  F.~Hutter.
\newblock {SMAC}v3: Algorithm configuration in {P}ython, 2017.

\bibitem[Schulman et~al.(2017)Schulman, Wolski, Dhariwal, Radford, and
  Klimov]{schulman-arxiv17a}
J.~Schulman, F.~Wolski, P.~Dhariwal, A.~Radford, and O.~Klimov.
\newblock Proximal policy optimization algorithms.
\newblock \emph{arXiv:1707.06347}, 2017.

\bibitem[Shahriari et~al.(2016)Shahriari, Swersky, Wang, Adams, and
  de~Freitas]{shahriari-ieee16a}
B.~Shahriari, K.~Swersky, Z.~Wang, R.~Adams, and N.~de~Freitas.
\newblock Taking the human out of the loop: {A} review of {B}ayesian
  optimization.
\newblock \emph{Procs. of the {IEEE}}, 104\penalty0 (1):\penalty0 148--175,
  2016.

\bibitem[Snoek et~al.(2012)Snoek, Larochelle, and Adams]{snoek-NeurIPS12a}
J.~Snoek, H.~Larochelle, and R.~Adams.
\newblock Practical {B}ayesian optimization of machine learning algorithms.
\newblock In \emph{Proc. of NeurIPS'12}, pages 2960--2968, 2012.

\end{thebibliography}

\end{document}